\pdfoutput=1
\documentclass[11pt]{article}
\usepackage{coling2020}
\usepackage{times}
\usepackage{url}
\usepackage{latexsym}

\usepackage{amsmath}
\usepackage{amsfonts}
\usepackage{microtype}
\usepackage{xspace}
\usepackage{graphicx} 
\usepackage{tabularx}
\usepackage{booktabs}
\usepackage{url}
\usepackage{relsize}
\usepackage{multirow}
\usepackage{paralist}
\usepackage{caption}
\usepackage{subcaption}

\newcommand{\sd}[1]{\smaller[2]\ensuremath{\pm#1}}
\newcommand{\nosd}{\ensuremath{^*}}
\newcommand{\todomain}{$\rightarrow$}

\newcommand{\ex}[1]{\textit{#1}\xspace}
\newcommand{\means}[1]{``#1''\xspace}

\newcommand{\class}[1]{\texttt{#1}\xspace}

\newcommand{\dataset}[1]{\textsc{#1}\xspace}
\newcommand{\relocar}{\dataset{ReLocaR}}
\newcommand{\semeval}{\dataset{SemEval}}
\newcommand{\semevalORG}{\dataset{SemEval$_{\text{ORG}}$}}
\newcommand{\semevalLOC}{\dataset{SemEval$_{\text{LOC}}$}}
\newcommand{\gwn}{\dataset{GWN}}
\newcommand{\conll}{\dataset{CoNLL}}
\newcommand{\wimcor}{\dataset{WiMCor}}

\newcommand{\tabref}[1]{Table~\ref{#1}\xspace}
\newcommand{\figref}[1]{Figure~\ref{#1}\xspace}

\newcommand{\secref}[1]{Section\xspace\ref{#1}\xspace}

\newcommand{\method}[2][]{\textsc{#2#1}\xspace}
\newcommand{\gyder}{\method{GYDER}}
\newcommand{\prewingv}{\method{PreWin$_{\text{GV}}$}}
\newcommand{\prewinbert}{\method{PreWin$_{\text{BERT}}$}}
\newcommand{\bertbase}[1][]{\method[#1]{BERT-base}}
\newcommand{\bertlarge}[1][]{\method[#1]{BERT-lg}}
\newcommand{\aug}{\method{+aug}}
\newcommand{\mask}{\method{+mask}}

\colingfinalcopy 

\title{Target Word Masking for Location Metonymy Resolution}

\author{Haonan Li$^\spadesuit$  \hspace{1cm} Maria Vasardani$^\diamondsuit$ \hspace{1cm} Martin Tomko$^\heartsuit$ \hspace{1cm} Timothy Baldwin$^\spadesuit$\\
	$\spadesuit$ School of Computing and Information Systems, The University of Melbourne  \\
	$\heartsuit$ Department of Infrastructure Engineering, The University of Melbourne \\
	$\diamondsuit$ Department of Geospatial Science, RMIT University \\
	\tt haonanl5@student.unimelb.edu.au, maria.vasardani2@rmit.edu.au\\ 
	\tt tomkom@unimelb.edu.au, tb@ldwin.net\\
} 

\date{}

\begin{document}

\maketitle

\begin{abstract}
  Existing metonymy resolution approaches rely on features extracted
  from external resources like dictionaries and hand-crafted lexical
  resources.  In this paper, we propose an end-to-end word-level
  classification approach based only on BERT, without dependencies on
  taggers, parsers, curated dictionaries of place names, or other
  external resources. We show that our approach achieves the
  state-of-the-art on 5 datasets, surpassing conventional BERT models
  and benchmarks by a large margin. We also show that our approach
  generalises well to unseen data.
\end{abstract}

\section{Introduction}

Metonymy is a widespread linguistic phenomenon, in which a thing or
concept is referred to by the name of something closely associated with
it. It is an instance of figurative language that can be easily
understand by humans through association, but is hard for machines to
interpret.  For example, in \ex{They read \underline{Shakespeare}}, it is \means{the
  works of Shakespeare} that we are referring to, not the playwright
himself.

Existing named entity recognition (NER) and word sense disambiguation
(WSD) systems have no explicit metonymy detection. This is an issue as
named entities and other lexical items are often used metonymically. For
instance, \ex{Germany} in the context of \ex{Germany lost in the
  semi-final} refers to \means{the national German sports team},
different to the context \ex{I live in Germany} which the term is used
literally. NER systems generally tag both as \class{location} without
recognition of the metonymic usage in the first, and WSD systems are
tied to word sense inventories and generally don't handle metonyms and
other sense extensions well \cite{DBLP:journals/lre/MarkertN09}. Intuitively, metonym resolution should
improve NER and WSD, something we explore in this paper.

Metonymy resolution is the task of determining whether a potentially
metonymic word (``PMW'') in a given context is used metonymically or
not. It has been shown to be an important component of many NLP tasks,
including machine translation \cite{kamei-wakao-1992-metonymy}, question
answering \cite{stallard-1993-two}, anaphora resolution
\cite{DBLP:journals/ai/MarkertH02}, geographical information retrieval
\cite{DBLP:journals/gis/LevelingH08}, and geo-tagging
\cite{DBLP:journals/gandc/MonteiroDF16,DBLP:journals/lre/GrittaPLC18}.

Conventional approaches to metonymy resolution have made extensive 
use of taggers, parsers, lexicons, and corpus-derived or hand-crafted 
features \cite{nissim-markert-2003-syntactic,farkas-etal-2007-gyder,nastase-etal-2012-local}. 
These either rely on NLP pre-processors that potentially 
introduce errors, or require external domain-specific resources. Recently, 
deep contextualised word embeddings \cite{peters-etal-2018-deep} and 
pre-trained language models \cite{devlin-etal-2019-bert} have been shown 
to benefit many NLP tasks, and part of our interest in this work is how to 
best apply these approaches to metonymy resolution.

While we include experiments for other types of metonymy, a particular
focus of this work is locative metonymy. Previous work has suggested
that around 13--20\% of toponyms are metonymic
\cite{markert-nissim-2007-semeval,DBLP:journals/gis/LevelingH08,Gritta2019},
such as in \ex{Vancouver welcomes you}, where \ex{Vancouver} refers to
\means{the people of Vancouver} rather than the literal place.

Our contributions are as follows. First, we propose a word masking
approach, which when paired with fine-tuned BERT
\cite{devlin-etal-2019-bert}, achieves state-of-the-art accuracy over a
number of benchmark metonymy datasets: our method outperforms the
previous state-of-the-art by 5.1\%, 12.2\% and 4.8\%, on \semeval
\cite{markert-nissim-2007-semeval}, \relocar
\cite{gritta-etal-2017-vancouver} and \wimcor \cite{kevin-2020-met},
respectively, and also outperforms a conventional fine-tuned BERT model
by a large margin.  Second, in addition to intrinsic evaluation of
location metonymy resolution, we include an extrinsic evaluation, where
we incorporate a locative metonymy resolver into a geoparser, and show
that it boosts geoparsing performance.  Third, we demonstrate that our
method generalises better cross-domain, while being more data efficient.
Finally, we conduct a detailed error analysis from the task rather than
model perspective. Our code is available at: \url{https://github.com/haonan-li/TWM-metonymy-resolution}.

\section{Related Work}

In early symbolic work, metonymy was treated as a syntactico-semantic
violation \cite{hobbs1987local,pustejovsky1991generative}.  As such, the
resolution of metonymy was based on constraint violation, usually based
on the selectional preferences of verbs
\cite{fass1991met,hobbs1993interpretation}.

\newcite{markert-nissim-2002-metonymy} were the first to treat metonymy
resolution as a classification task, based on corpus and linguistic
analysis. They demonstrated that grammatical roles and syntactic
associations are high-utility features, which they subsequently extended
to include syntactic head--modifier relations and grammatical roles
\cite{nissim-markert-2003-syntactic}. To tackle data sparseness, they
further introduce simpler grammatical features by integrating a
thesaurus.  Much of this work has been preserved in recent work in the
form of hand-engineered features and external resources.

SemEval 2007 Task 8 on metonymy resolution
\cite{markert-nissim-2007-semeval} further catalyzed interest in the
task by releasing a metonymy dataset with syntatic and grammatical
annotations, and fine-tuning the task definition and evaluation metrics.
A range of learning paradigms (including maximum entropy, decision
trees, and naive Bayes) were applied to the task. Top-ranking systems
\cite{nicolae-etal-2007-utd,farkas-etal-2007-gyder,brun-etal-2007-xrce}
used features provided by the organisers, such as syntactic roles and
morphological features. Most systems also used features from external
resources such as WordNet, FrameNet, VerbNet, and the British National
Corpus (BNC).

Later work
\cite{nastase-strube-2009-combining,nastase-etal-2012-local,DBLP:journals/ai/NastaseS13}
used the Wikipedia category network to capture the global context of
PMWs, to complement local context features. 

All the above-mentioned approaches resolve metonymy by enriching the
information about PMWs, in particular via resources. In contrast, our approach is end-to-end: information is contained in the pretrained
embeddings and language models only. Another difference is that we focus on the
context of the PMW only, and not the PMW itself.

More recently, in a departure from using ever-more hand-crafted
features, \newcite{gritta-etal-2017-vancouver} proposed a metonymy
resolution approach based on basic parsing features and word embeddings.
The main idea is to eliminate words that are superfluous to the task and
keep only relevant words, by constructing a ``predicate window'' from
the target word via a syntactic dependency graph.  The classification of
the target word is then based on the ``predicate window''. Similar to
us, they do not take the identity of the target word into
consideration. However, we remove the dependency on a dependency parser,
and more systematically generate a context representation by masking
the target word within a pretrained language model.

Researchers have released several datasets for metonymy resolution,
including \semeval \cite{nissim-markert-2003-syntactic}, \relocar and
\conll \cite{gritta-etal-2017-vancouver}, \gwn \cite{Gritta2019}, and
\wimcor \cite{kevin-2020-met}. However, none of them have analyzed the
data distribution and or generalisation across datasets. In this paper,
we train our model on different datasets, and evaluate its transfer
learning abilities.

\section{Approach}

Formally, given a sentence and target word\footnote{In practice, the
  target ``word'' can be made up of multiple tokens, either due to it
  containing multiple word pieces, or being a longer phrase such as
  \ex{New York}.} contained within it, metonymy resolution is the
classification of whether the target word is a metonym or not, and what
metonymic readings it has.

\subsection{Motivation}

Due to the relatively small size of most existing metonymy resolution
datasets, researchers have explored ways to compensate for the sparse
training data, e.g.\ through data augmentation
\cite{kobayashi-2018-contextual,wei-zou-2019-eda}.  Modern pre-trained
language models offer an alternative approach which performs well when
fine-tuned over even small, task-specific datasets
\cite{DBLP:conf/icml/HoulsbyGJMLGAG19,porada-etal-2019-gorilla}.  Data 
sparseness may, however, still lead to overfitting. For example, in our metonymy resolution task, 
if the target word \ex{Vancouver} appears only once during training, in
the form of a metonymy, the model might overfit and always predict that
\ex{Vancouver} is a metonymy regardless of context. Intuitively, masking
the target word during training can eliminate lexical bias and force the
model to learn to classify based on the context of use rather than the
target word.

\subsection{BERT for Word-level Classification}

For the tokenised sentence $S=\langle t_1,t_2,...t_n \rangle$ and target
word $\langle t_i,...t_j \rangle$ with position pair $(i,j)$, we form
the input sentence to the BERT encoder: \texttt{[CLS] + $S$ + [SEP]}. We
extract the representation of the target word from the last hidden
layer as $\mathbf{T}\in\mathbb{R}^{d \times h}$, where $d=j-i+1$ denotes
the length of the target word and $h$ is the hidden layer
size. Element-wise averaging is applied to the word span, such that the
extracted matrix is compressed into the vector
$\mathbf{T}\in\mathbb{R}^{1 \times h}$. Finally, we feed it into a
linear classifier and get the output as the label.

\subsection{Data Augmentation}
\label{sec:data-aug}

One method to combat lexical overfitting for the target word due to the
small data setting is data augmentation. We first extract all target
words from the training set as a target word pool. Then, for each
training sample, a fresh sample is constructed by replacing the target
word with a random one from the pre-built pool. We repeat this 10 times,
expanding the training data set 10-fold in the process. We train the
models on the augmented training set and evaluate on the original test
set.

\subsection{Target Word Masking}
\label{sec:word-mask}

An alternative approach is to mask the target word, and base the
classification exclusively on the context. We claim that the
interpretation (metonym or literal) of a target word relies more on the
context of use than the word itself. To test this claim, we force the
model to predict whether the target word is metonymic based only on
context. Here, we replace the input target word with the single token
\ex{X} during training and evaluation. Note that this is not compatible
with the data augmentation method described in \secref{sec:data-aug}, as the target word
(either original or replaced through data augmentation) is masked out.

\subsection{End-to-end Metonymy Resolution}

Given a raw sentence, an end-to-end metonymy resolution requires that
the model can detect PMWs and predict the correct class of each. Most
existing metonymy resolution methods focus on named entities (e.g.\
locations and organizations), which we detect by training a BERT named
entity recogniser to detect locations and organizations based on the
CoNLL 2003 data. These detected locations and organizations are masked
one at a time, and fed into the word-level BERT classifier.

\section{Experimental Details}

In this section, we detail the five datasets used in our experiments,
and then provide details of the models used in this research. 

\subsection{Datasets}\label{sec:dataset}

\begin{table}[t!]
	\begin{center}
		\begin{tabular}{lrrrrcc}
			\toprule 			
			\textbf{Dataset}& $\#$literal &$\#$metonym & Total &$\#$PMWs & Avg. doc. length & Source\\
			\midrule
			\semevalORG 	& 1211 & 721 & 1932 & 433 & 27.3 & BNC\\
			\semevalLOC	 	& 1458 & 375 & 1833 & 262 & 26.6 & BNC\\
			\relocar		 	& 995 & 1031 & 2026 & 603 & 22.7 & Wikipedia\\
			\conll 				& 4609 & 2448 & 7057 & 1685 & 24.6 & CoNLL\\
			\gwn 				& 841 & 630 & 1471 & 600 & 26.8 & News\\
			\wimcor	 		& 154322 & 51678 & 206000 &1029 & 85.3& Wikipedia\\
			\bottomrule
		\end{tabular}
	\end{center}
	\caption{Statistics of the datasets used in this research. The document length is in terms of the number of words}
	\label{dataset}
\end{table}

\textbf{\semeval} was first introduced by
\newcite{nissim-markert-2003-syntactic} and subsequently used in SemEval
2007 Task 8 \cite{markert-nissim-2007-semeval}. It contains about 3800
sentences from the BNC across two types of entities: organizations and
locations. In addition to coarse-level labels of metonym or literal, it
contains finer-grained labels of metonymic patterns, such as
place-for-people, place-for-event, or place-for-product. This is the
only dataset where have such fine-grained labels of metonymy. We use the
dataset in two forms, based on the entity type of the target word: (1)
spatial metonymies (``\semevalLOC''); and organization metonymies
(``\semevalORG'').

\textbf{\relocar} \cite{gritta-etal-2017-vancouver} is a Wikipedia-based
dataset. Compared with \semevalLOC, it is intended to have better label
balance,
and annotation quality, without the fine-grained analysis of metonymic
patterns. It contains 2026 sentences, and is focused on locations
only. It is important to note that the class definitions for \relocar
are a bit different from those for \semevalLOC. The main difference is in
the interpretation of \class{political entity} (e.g.\ \ex{Moscow
  opposed the sanctions}), which is considered to be a literal reading in
\semeval, but metonymic in \relocar. The argument is that 
governments/nations/political entities (in the case of our example,
\means{the government of Russia}) are much closer to organizations or
people semantically, and thus metonymic.

\textbf{\conll} was released together with \relocar
\cite{gritta-etal-2017-vancouver} and also focused on locations. It contains about 7000 sentences
taken from the CoNLL 2003 Shared Task on NER and was annotated by one
annotator only, with no quantification of the quality of the labels, and
is thus potentially noisy.

\textbf{\gwn} \cite{Gritta2019} is a fine-grained labelled dataset of
toponyms consisting of around 4000 sentences. It contains not only
metonymic usages of locations, but also demonyms,\footnote{A demonym is a word that identifies a group of people (inhabitants, residents, natives) in relation to a particular place.} homonyms, and noun
modifiers, of which we extract instances labelled as \class{literal},
\class{metonymic}, or \class{mixed} in our experiments. We merge the
\class{mixed} instances (which account for around 2\% of the data) with
\class{metonymy}, creating a binary classification task.

\textbf{\wimcor} \cite{kevin-2020-met} is a semi-automatically annotated
dataset from English Wikipedia, based on the observation that Wikipedia
disambiguation pages list different senses of ambiguous entities. The
authors use disambiguation pages to identify literal and metonymic
entities, and extract Wikipedia article pairs with the same natural
title which refer to different but related concepts, like \ex{Delft} and
\ex{Delft University of Technology}. Sentences are then extracted from
the backlinks of the respective articles, which point to the articles
that contain the target mentions.  The dataset contains 206K samples, of
which about one-third are metonyms.  Although the dataset is
large-scale, it contains only 1029 unique PMWs, which means that in the
standard data split there are few unseen PMWs in the test data. To make
the task more difficult, and avoid possible lexical memorization
\cite{levy-etal-2015-supervised,Vylomova+:2016a}, we employ a different split, to ensure no PMWs occur in both
the training and test splits.

A statistical breakdown of the five datasets is provided in
\tabref{dataset} (noting that \semevalLOC and \semevalORG are listed
separately, making a total of six listings for our five datasets). Note
that all datasets are in English.

\subsection{Baselines}

The first two baselines are the best model \cite{farkas-etal-2007-gyder} on SemEval-2007 Shared Task, \gyder, and results reported by \newcite{nastase-strube-2009-combining} and \newcite{nastase-etal-2012-local} in following years. We simply report their results without reimplementing the models.
We reimplement the SOTA PreWin model \cite{gritta-etal-2017-vancouver}
as another baseline.  To do this, we first generate the dependency structures
and labels using SpaCy,\footnote{\url{https://spacy.io/}} and index the
predicate window by the dependency head of the PMW. The output of the
predicate window is then fed into two LSTM layers, one for the left
context and one for the right context. The dependency relation labels of
the content of the predicate window are represented as one-hot vectors
and feed into two dense layers, for the left and right contexts,
separately. By concatenating the four layers' output and feeding it to a
multi-layer perceptron, we get the final label. In line with the
original paper, we use GloVe embeddings to represent the words
\cite{Pennington+:2014}, set the window size to 5, use a dropout rate of
0.2, and trained the model for 5 epochs.

To make the baseline model more competitive with our approach, we
additionally experiment with a variant of the baseline where we replace
the original GloVe embeddings with BERT embeddings
\cite{devlin-etal-2019-bert}. We experimented with both BERT-base and
BERT-large, but present results for BERT-base as we observed no
improvement using the larger model.

\subsection{Our Model}

We use BERT in three settings, for both the BERT-base (``\bertbase'')
and BERT-large (``\bertlarge'') models: (1)
fine-tuned over a given dataset, with no masking; (2) fine-tuned with
data augmentation (see \secref{sec:data-aug}); and (3) fine-tuned using
target word masking (see \secref{sec:word-mask}).  We use the uncased
model with a learning rate of 5e-5, and max sequence length of 256.  For
\wimcor, we fine-tune for 1 epoch with a batch size of 64, and dropout rate of
0.2. For the other datasets, we fine-tune for 10 epochs with a batch size of 64
and dropout rate of 0.1.\footnote{All hyper-parameter tuning was based
  on development data.}

\subsection{BERT Ensemble}

Due to the large number of parameters in BERT and small size of the
training datasets (with the exception of \wimcor), the models tend to
overfit or be impacted by bad initialisations. To counter this, we
experimented with ensembling different runs of a given BERT model,
specifically, the BERT-large model with word masking.

\subsection{Cross-domain Transfer}

In \secref{sec:dataset} we noticed that the different metonymy datasets
were created with different annotation guidelines and over different
data sources. To study the ability of the different models to generalise
across datasets, we train models on one dataset and evaluate on a
second, in the following six configurations: train on \semevalLOC and
test on \relocar (and vice versa); and train on \conll and \wimcor
separately and test on either \semevalLOC or \relocar. For all 6
settings, we compare the PreWin model of
\newcite{gritta-etal-2017-vancouver} with the three BERT settings
(basic; with data augmentation; and with target word masking), all based
on the BERT-large-uncased model.

\subsection{Extrinsic Evaluation}

To extrinsically evaluate our proposed method, we combine different
metonymy resolution methods with a state-of-the-art geoparser, and
evaluate over the \gwn dataset. 
The task is to detect the locations with literal reading only and ignore all other possible readings.
Following \newcite{Gritta2019}, we classify toponyms as either \class{literal} or
\class{associative}.\footnote{Associative toponyms include metonyms,
  homonyms, demonyms, and noun modifiers.} We simply pipeline the
Edinburgh Geoparser (without fine-tuning) \cite{grover2010use} with our metonymy resolver as a
baseline. The Edinburgh Geoparser detects all toponyms through an NER sequence supported by the Geonames gazetteer, but does not indicate metonymic usage. After this toponym detection, our metonymy resolver filters out non-literal uses of toponyms. The other two baselines used here are a reimplementation of the NCRF++
tagger of \newcite{Gritta2019},\footnote{Note that we were not able to
  fully reproduce the published results: the F-score in the original
  paper is 77.6, as compared to 77.2 for our reimplementation.}  and
\bertlarge fine-tuned on the geoparsing task. For our end-to-end model,
we separate geoparsing into the toponym detection and metonymy
resolution subtasks, and fine-tune the NER part on toponym detection,
and the masked model on metonymy resolution.

\begin{table}[t!]
	\begin{center}
		\begin{tabular}{lr@{}lr@{}lr@{}lr@{}lr@{}l}
			\toprule
			\textbf{Method} & \multicolumn{2}{c}{\semevalLOC} & \multicolumn{2}{c}{\relocar} & \multicolumn{2}{c}{\conll} & \multicolumn{2}{c}{\gwn} & \multicolumn{2}{c}{\wimcor} \\
			\midrule
			\gyder              & 85.2          & \nosd        & --- &          & --- &                  & --- &           & --- \\
			Nastase et al.      & {86.2}        & \nosd      & --- &           & --- &                & --- &            & --- \\
			\prewingv           & 83.1          & \sd{0.64}  & {83.6}     & \sd{0.71}   & {87.9}     & \sd{0.22}   & 83.4 & \sd{1.24} & 90.7 & \sd{0.15} \\
			\prewinbert         & 87.1          & \sd{0.54}  & {92.2}     & \sd{0.48}   & {92.6}     & \sd{0.32}   & 89.1 & \sd{0.96} & 93.7 & \sd{0.07} \\
			\midrule
			\bertbase           & 85.0          & \sd{0.46}  & 91.5       & \sd{0.54}   & {90.9}     & \sd{0.31}   & 86.8      & \sd{1.00} & 93.6 & \sd{0.12} \\
			\bertbase[\aug]     & 84.5          & \sd{0.85}  & 91.0       & \sd{0.72}   & ---     &   & 85.2      & \sd{0.76} & --- & \\
			\bertbase[\mask]    & {87.1}        & \sd{0.89}  & {93.9}     & \sd{0.52}   & {93.7}     & \sd{0.46}   & {89.5}    & \sd{0.47} & 95.4 & \sd{0.10} \\
			\midrule
			\bertlarge          & 84.7          & \sd{0.71}   & 91.3       & \sd{0.57}   & {89.5}     & \sd{0.84}   & 88.3      & \sd{1.02} & 93.7 & \sd{0.17} \\
			\bertlarge[\aug]    & 85.0          & \sd{1.10}  & 91.4       & \sd{0.86}   & ---     &     & 86.1      & \sd{1.21} & --- & \\
			\bertlarge[\mask]   & \bf{88.2}     & \sd{0.61}   & \bf{94.4}  & \sd{0.31}   & \bf{93.9}     & \sd{0.54}   & \bf{91.2} & \sd{0.40} & \bf{95.5} & \sd{0.13} \\
			\midrule
			Ensembled \bertlarge[\mask] 		& \bf{89.1} &				&	\bf{94.8} 	&					& \bf{94.6}     &       &\bf{92.0}&					&\bf{95.9}&\\
			\bottomrule
		\end{tabular}
	\end{center}
	\caption{Accuracy (\%) of metonymy resolution for the locative
          datasets, averaged over 10 runs with standard deviation; the
          best result for each dataset is indicated in boldface, and
          ``\nosd'' denotes the result published by the authors.}
	\label{res:loc}
\end{table}

\begin{table}[t!]
	\begin{center}
		\begin{tabular}{lr@{}lr@{}lr@{}lr@{}lr@{}l}
			\toprule
			\textbf{Method} & \multicolumn{2}{c}{\semevalORG}  \\
			\midrule
			\gyder               & 76.7      & \nosd  \\
			Nastase et al.      & {77.0}    & \nosd \\
			\prewingv           & 73.0     & \sd{0.72}  \\
			\prewinbert         & \bf{79.6}  & \sd{0.48} \\
			\midrule
			\bertbase           & 72.4      & \sd{1.03} \\
			\bertbase[\aug]       & 73.5      & \sd{1.38}  \\
			\bertbase[\mask]      & {75.6}    & \sd{1.47} \\
			\midrule
			\bertlarge          & 74.3      & \sd{1.12}  \\
			\bertlarge[\aug]      & 75.1      & \sd{0.58} \\
			\bertlarge[\mask]     & 77.2 & \sd{1.15}  \\
			\midrule
                  Ensembled \bertlarge[\mask]       &\bf{79.6}&	\\
			\bottomrule
		\end{tabular}
	\end{center}
	\caption{Accuracy (\%) of metonymy resolution for the
          non-locative dataset, averaged over 10 runs with standard
          deviation; the best result is indicated in boldface, and
          ``\nosd'' denotes the result published by the authors.}
	\label{res:other}
\end{table}

\section{Results}

To evaluate metonymy resolution, we train each model over 10 runs and
report the average accuracy and standard deviation. For geoparsing, we
use precision, recall, and F1-score, based on 5-fold cross-validation.

\tabref{res:loc} shows the results of metonymy resolution across the
five locative datasets. For all datasets, both \bertbase and \bertlarge
outperform the previously best-published results. The use of BERT in
\prewinbert clearly improves the method over the original \prewingv,
but the results are consistently lower than \bertlarge[\mask].  Data
augmentation (``aug'') sometimes improves and sometimes degrades
performance, whereas target word masking (``mask'') consistently
improves performance. The best single model results for all datasets are
achieved with our \bertlarge[\mask] model, and the ensemble version
improves results a bit further. We didn't apply data augmentation on
\conll and \wimcor because the two datasets are sufficiently large
without it.

Comparing the different datasets, the relative accuracies vary
substantially: \semevalLOC is the most difficult, while \wimcor is the
simplest, even with the lexically-split training and test data. With the
original data split for \wimcor, the result is over 99.5 even with \bertbase without
masking (and even higher for the other BERT-based methods).

Although it is not the main focus of this paper, we also report the
results for the non-locative dataset in \tabref{res:other}. Once again,
our masked model attains consistent improvements over the unmasked
model. The best result is achieved by \prewinbert and the ensembled
version of \bertlarge[\mask].

\tabref{res2} shows the results of the cross-domain experiments. From
the first two rows we see that generalization between \semevalLOC and
\relocar is not good, which we hypothesise is due to the differences in
the annotation schemes and label distributions. In contrast, models
trained on \conll transfer better to \relocar and \semevalLOC. The last
two rows show the cross-domain results from \wimcor, where despite
\wimcor containing orders of magnitude more data than the other
datasets, it transfers poorly to both \relocar and \semevalLOC. This
supports our conjecture that \wimcor is more one-dimensional than the
other datasets, making it hard to generalise, even with the additional
training data. Overall, the models using either BERT embedding or
fine-tuned BERT perform better than PreWin with GLOVE embedding, and
our masking approach always give the models better generalisation
ability.

\tabref{res3} shows the geoparsing results on \gwn. The Edinburgh
Geoparser does not perform well, as it is not fine-tuned to the dataset.
The BERT tagger outperforms NCRF++, and our end-to-end model beats
BERT. This is evidence that incorporating explicit metonymy resolution
into a geoparser improves its performance, and also that our metonymy
resolution method is sufficiently accurate to improve over a comparable
model without metonymy resolution.

\begin{table}[t!]
	\begin{center}
          \resizebox{\textwidth}{!}{
          \begin{tabular}{r@{}c@{}lr@{}lr@{}lr@{}lr@{}lr@{}l}
			\toprule
			\multicolumn{3}{c}{\multirow{2}{*}{\textbf{Source \todomain Target}}} & \multicolumn{10}{c}{\textbf{Method}}\\
			\cmidrule{4-13}
			&&& \multicolumn{2}{c}{\prewingv} & \multicolumn{2}{c}{\prewinbert} & \multicolumn{2}{c}{\bertlarge} & \multicolumn{2}{c}{\bertlarge[\aug]} & \multicolumn{2}{c}{\bertlarge[\mask]} \\
			\midrule
			\semevalLOC&\todomain&\relocar  & 62.4 & \sd{2.30} & 74.0 & \sd{1.65} & 65.9 & \sd{1.81} & 66.8 & \sd{2.50} & \textbf{75.2} & \sd{1.05}\\
			\relocar&\todomain&\semevalLOC  & 69.0 & \sd{3.13} & 72.7 & \sd{1.08} & 71.9 & \sd{2.63} & 70.0 & \sd{1.79} & \textbf{74.8} & \sd{1.29}\\
			\conll&\todomain&\relocar       	  & 82.6 & \sd{0.87} & 90.2 & \sd{0.70} & 88.9 & \sd{0.69} & 88.1 & \sd{0.66} & \textbf{93.5} & \sd{0.40}\\
			\conll&\todomain&\semevalLOC    & 79.5 & \sd{0.34} &80.7  & \sd{0.73} & 81.0 & \sd{1.16} & 80.8 & \sd{0.79} & \textbf{82.5} & \sd{1.69}\\
			\wimcor&\todomain&\relocar          &  64.1  & \sd{0.42} &  60.4 & \sd{0.60} & 51.6 & \sd{1.55} & --- &  & \textbf{64.6} & \sd{1.05}\\
			\wimcor&\todomain&\semevalLOC  &   75.6   & \sd{0.76} & 78.2  & \sd{0.94} & 78.2 & \sd{0.56} & --- &  & \textbf{78.4} & \sd{0.97}\\
			\bottomrule
		\end{tabular}}
	\end{center}
	\caption{Cross-domain accuracy (averaged over 10 runs, with standard deviation).}
	\label{res2}
\end{table}

\begin{table}[t!]
	\begin{center}
		\begin{tabular}{lr@{}lr@{}lr@{}l}
			\toprule
			\textbf{Model} & \multicolumn{2}{c}{\textbf{Precision}} & \multicolumn{2}{c}{\textbf{Recall}} & \multicolumn{2}{c}{\textbf{F-score}} \\
			\midrule
			E-GeoParser & 76.0          & \sd{2.10} & 48.6          & \sd{4.38} & 59.5          & \sd{3.61}\\ 
			NCRF++              & 79.5          & \sd{1.81} & 74.7          & \sd{1.09} & 77.1          & \sd{1.13}\\
			BERT tagger         & 80.1          & \sd{2.38} & 81.0          & \sd{1.00} & 80.5          & \sd{1.53}\\
			Our method          & \textbf{80.9} & \sd{1.58} & \textbf{81.3} & \sd{1.19} & \textbf{81.1} & \sd{0.93}\\
			\bottomrule
		\end{tabular}
	\end{center}
	\caption{Geoparsing results (averaged over 5 folds of cross validation, with standard deviation). E-Geoparser is the Edinburgh Geoparser.}
	\label{res3}
\end{table}

\begin{figure}[t!]
	\begin{subfigure}{.5\textwidth}
		\centering
		\includegraphics[width=1\linewidth]{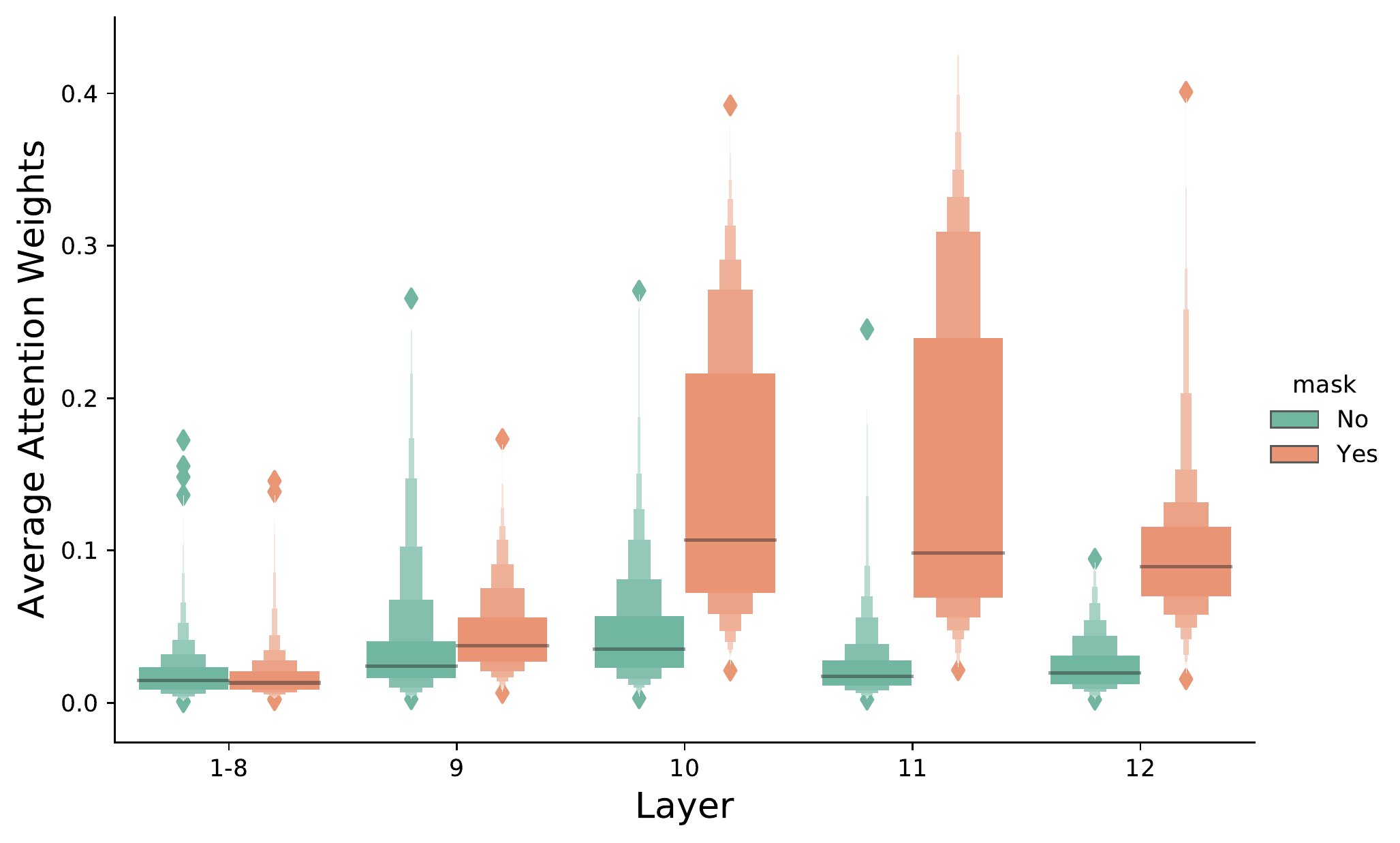}
		\caption{\bertbase model }
		\label{fig:weights_base}
	\end{subfigure}
	\begin{subfigure}{.5\textwidth}
		\includegraphics[width=1\linewidth]{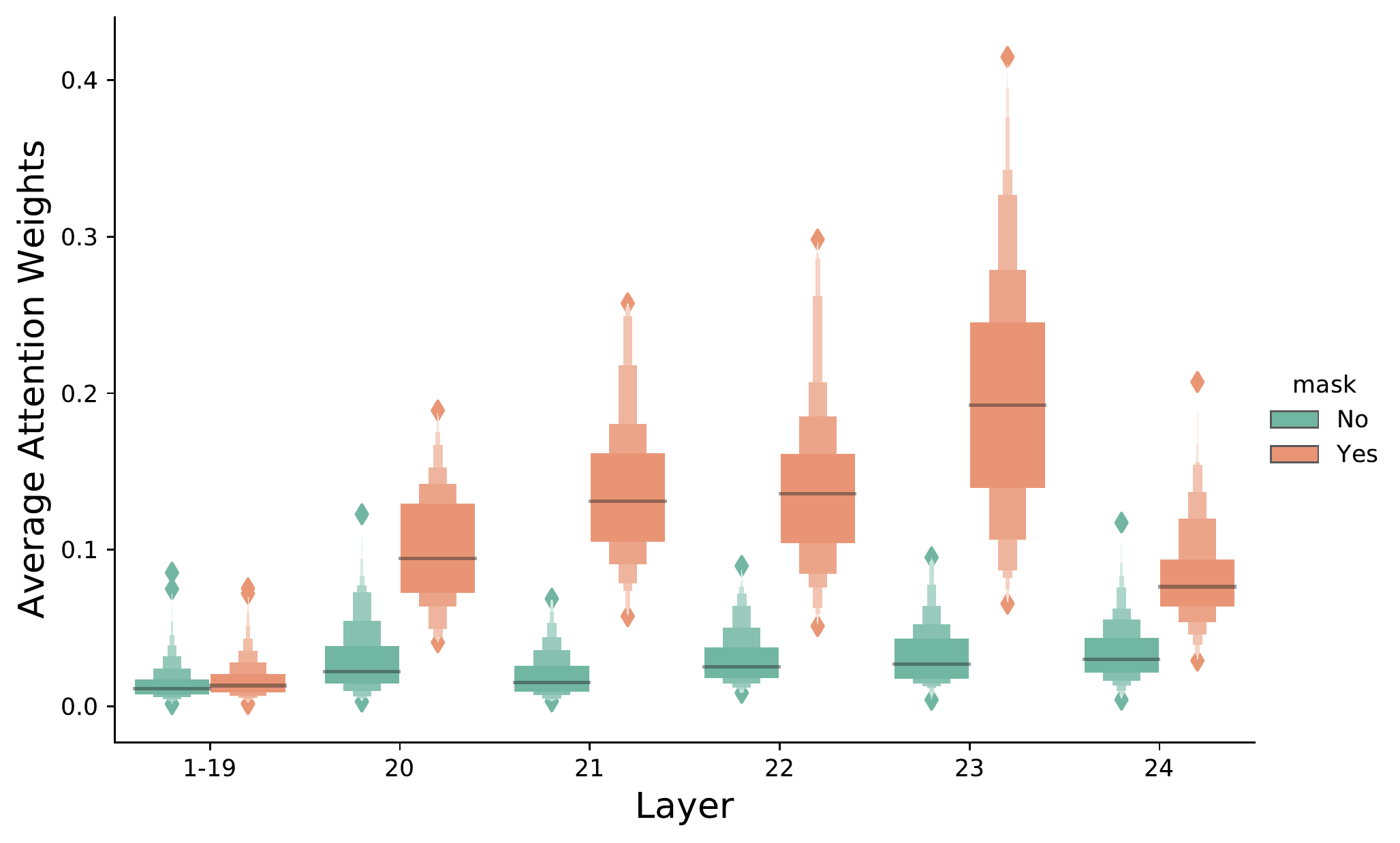}
		\caption{\bertlarge model }
		\label{fig:weights_large}
	\end{subfigure}
	\caption{Visualisation of the self-attention weights on the target
		word on the \gwn dataset. Layers 1--8  and 1--19 are merged for \bertbase[\mask] and
		\bertlarge[\mask], respectively, as there is little difference between them. }
	\label{fig:weights}
\end{figure}

We further analysed the attention weights of the different fine-tuned
BERT models with and without target word masking. We compare the
attention weights for each layer separately (12 vs.\ 24 for \bertbase
and \bertlarge, resp.): we get the attention weight of each head on the
target word, and average the heads' weights to generate a single sample
point.  We found for both models, attention on the target word is
substantially higher for the last 4--5 layers, as shown in
\figref{fig:weights}. Moreover, target word masking makes the model
attend more to the target word.

\begin{figure}[t!]
	\centering
	\includegraphics[width=0.6\linewidth]{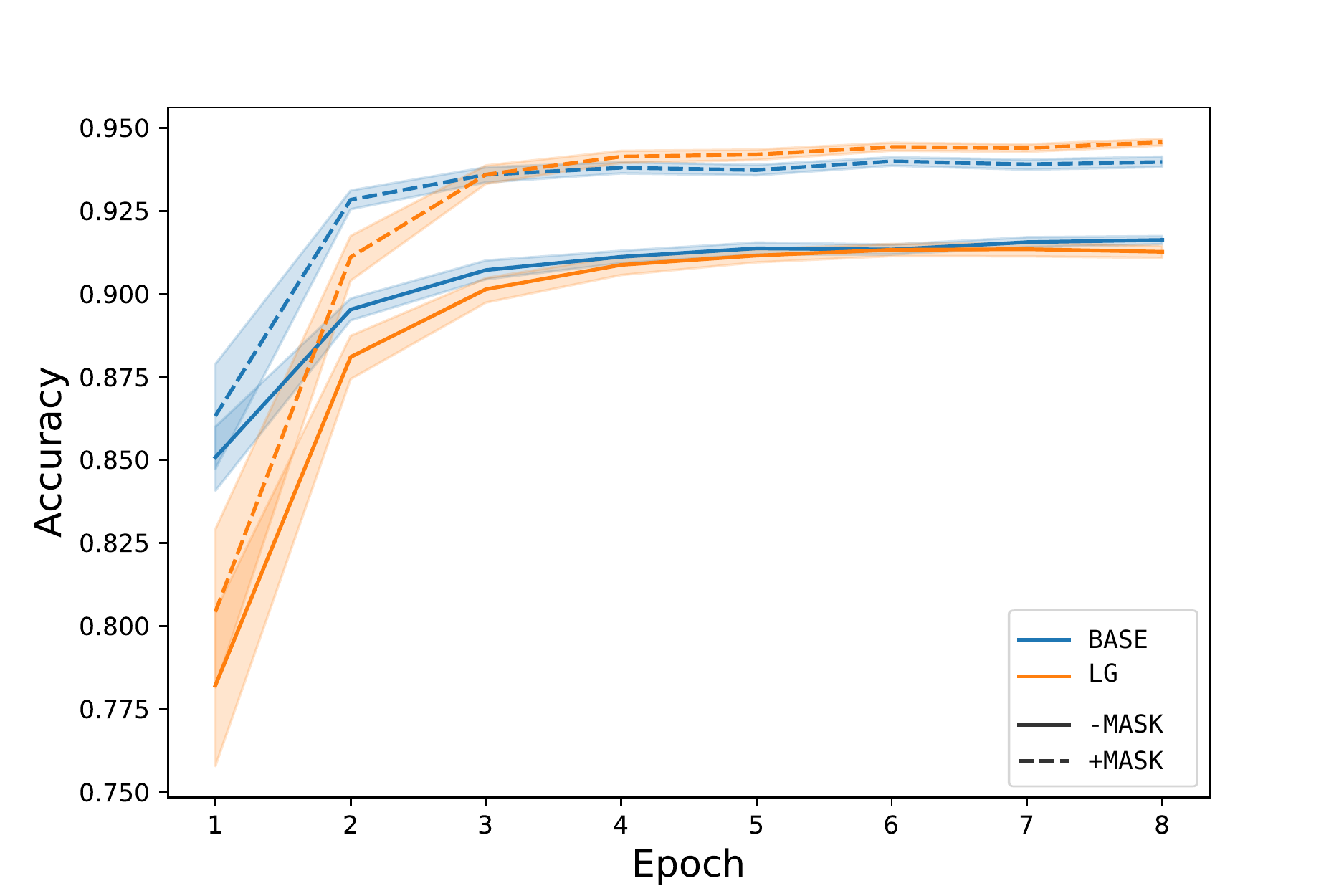}
	\caption{Training curve for the BERT models over \relocar. We
          show the averaged results for ten runs in each setting, with
          the shading indicating variance.}
	\label{fig:train_curve}
\end{figure}

\figref{fig:train_curve} is the training curve for the BERT models over
\relocar. We find that, generally, \bertlarge converges a bit slower than
\bertbase, but in each case, the masked model performs substantially
better than the unmasked models, and is more data efficient. While we
do not include them in the paper, the
plots for other datasets showed a similar trend.

\section{Error Analysis}

To further understand the task of metonymy resolution and why the model
fails in some cases, we conducted a manual error analysis over a random
sample of 150 errors from \semevalLOC and \relocar.  We roughly
categorise the errors into 6 types, with each instance classified
according to a unique error type.  Some instances had 
multiple errors among types 4, 5, and 6, in which case we classified them in
the priority of $6>5>4$.

\begin{enumerate}
\item \textbf{Data quality:}  Although the two datasets are labeled by experts,
  the inter-annotator agreement is not perfect, and the annotation
  guidlines are not exactly the same. For example,
  location-for-government is literal in \semevalLOC but metonymic in
  \relocar. Even based on the annotation guidelines used to generate the
  datasets, there are some labels that we do not agree with, such as the example
  for Type 1 in \tabref{error_analysis}: in our judgement, \ex{England} is the
  literal place or country here, but the gold label is metonym.
\item \textbf{Insufficient context:} Due to model capacity and
  efficiency reasons, we restrict the context to the context sentence of
  the PMWs. This removes useful context in some cases, such as the
  example for Type 2 in \tabref{error_analysis}, where the preceding
  sentence is: \ex{He will forsake China production schedules for fine
    tuning the first of many travel itineraries planned for ``my new
    career of retirement'', as he put it.} With this context, it is easier to recognise
  the PMW \ex{Canada} as an event, and hence a metonym. These errors
  can be resolved by including more context.
\item \textbf{Mixed meaning:} Some PMWs have both a literal and
  metonymic reading. We follow \newcite{gritta-etal-2017-vancouver} in
  treating them as metonymies, but the dominant reading is sometimes
  literal. In the example for Type 3 in \tabref{error_analysis},
  \ex{Malaysia} can either be the geographic place or the government of
  Malaysia. Such errors can be better handled with a more fine-grained
  classification schema.
\item \textbf{Long distance implications or complex syntactic
    structure:} The models struggle when the sentence structure becomes
  complex. For example, in the example for Type 4 in
  \tabref{error_analysis}, the immediate context suggests the PMW is
  literal, but the broader context suggests the opposite. Such issues
  can be addressed by using models with richer representations of
  sentence structure.
\item \textbf{Misleading wording and complex sentence semantics:} Complex
  semantics and grammatical phenomena like noun possessives confuse the
  model. \ex{America} in the phrase \ex{America's nuclear stockpile} has
  the literal reading, while \ex{Lebanon} in the phrase \ex{Lebanon's
    long ordeal} has a metonymic reading. A particularly subtle example
  is that for Type 5 in \tabref{error_analysis}, which requires the
  model to have near-human comprehension of semantics.
\item \textbf{Missing world knowledge:} Some examples require background
  knowledge to be understood, such as the example for Type 6 in
  \tabref{error_analysis}, where \ex{Vietnam} refers to an event (a war
  that happened there, and Bill Clinton's actions related to that war\footnote{\url{https://en.wikipedia.org/wiki/Bill_Clinton#Vietnam_War_opposition_and_draft_controversy}}). To deal with this, the model needs to have
  access to world knowledge, either implicitly or explicitly.
\end{enumerate}

These 6 error types vary in difficulty. From \tabref{error_analysis}, we
see that 62\% of current errors are caused by Types 5--6, namely the
model lacking an understanding of complex sentence semantics or world
knowledge, which are hard to solve. Possibly the only case with a clear
resolution is Type 2, where larger-context models may perform better.

\begin{table*}[t!]
	\begin{center}
		\begin{tabularx}{\textwidth}{cXlll}
			\toprule
			Type & Example & Label & Rate  \\
			\midrule
			1 & \ex{Jordan Houghton is a youth football player who plays in the \textbf{England} national under-17 football team.} & metonymy & 5\% \\
			2 & \ex{\textbf{Canada} in the Spring is already planned and Roy and his wife, Ann, are finalising the details of what promises to be an exciting six-week sojourn} & metonymy & 12\%  \\
			3 & \ex{The role of this office is to supply and coordinate aid from \textbf{Malaysia} to Kosovo and also to enable communication between the Malaysian government and the UNMIK.} & metonymy & 8\% \\
			4 & \ex{\textbf{Western Australia} becomes the last Australian state to abolish capital punishment for ordinary crimes (i.e., murder).} & metonymy & 8\%  \\
			5 & \ex{Yet, although \textbf{Britain} suffered severe economic depression and rising unemployment, her economic plight was much less marked than that of Germany and Italy.} & literal  &44\%  \\
			6 & \ex{Sex, drugs and \textbf{Vietnam} have haunted Bill Clinton's campaign for the US presidency.} & metonymy & 23\%  \\
			\bottomrule
		\end{tabularx}
	\end{center}
	\caption{Error analysis. ``Label'' is the gold label for the
          example, which our model was not able to predict.}
	\label{error_analysis}
\end{table*}

\section{Conclusions and Future Work}

In this paper, we proposed a word masking approach to metonymy
resolution based on pre-trained BERT, which substantially outperforms
existing methods over a broad range of datasets.  We also evaluated the
ability of different models in a cross-domain setting, and showed our
proposed method to generalise the best. We further demonstrated that an
end-to-end metonymy resolution model can improve the performance of a
downstream geoparsing task, and conducted a systematic error analysis of
our model.

The proposed target word masking method can be applied to tasks beyond
metonymy resolution. Numerous word-level classification tasks lack
large-scale, high-quality, balanced datasets. We plan to apply the
proposed word masking approach to these tasks to investigate whether it
can lead to similar gains over other tasks.

\section{Acknowledgements}

This research was sponsored in part by the Australian Research
Council. The authors would like to thank the anonymous reviewers for
their insightful comments.

\bibliographystyle{acl}
\bibliography{coling2020}

\end{document}